\documentclass[10pt,twocolumn,letterpaper]{article}

\usepackage[pagenumbers]{cvpr} 

\usepackage{graphicx}
\usepackage{amsmath}
\usepackage{amssymb}
\usepackage{booktabs}

\usepackage[pagebackref,breaklinks,colorlinks]{hyperref}

\usepackage[capitalize]{cleveref}
\crefname{section}{Sec.}{Secs.}
\Crefname{section}{Section}{Sections}
\Crefname{table}{Table}{Tables}
\crefname{table}{Tab.}{Tabs.}

\def\cvprPaperID{JoyBoy Team} 
\def\confName{CVPR}
\def\confYear{2022}

\begin{document}

\title{1st Place Solutions for RxR-Habitat Vision-and-Language Navigation Competition (CVPR 2022)}

\author{
\vspace{2mm}
Dong An$^{1,3*}$ \quad Zun Wang$^{2,4*}$ \quad Yangguang Li$^{3}$ \quad Yi Wang$^2$ \quad Yicong Hong$^4$ \\
\vspace{2mm}
\quad Yan Huang$^1$ \quad Liang Wang$^1$ \quad Jing Shao$^3$\\
$^1$Institute of Automation, CAS, $^2$Shanghai AI Laboratory, $^3$SenseTime\\ 
$^4$The Australian National University\\
{\tt\small andong2019@ia.ac.cn, zun.wang@anu.edu.au, liyangguang@sensetime.com, wangyi@pjlab.org.cn} \\
}

\maketitle

\begin{abstract}
This report presents the methods of the winning entry of the RxR-Habitat Competition in CVPR 2022. The competition addresses the problem of Vision-and-Language Navigation in Continuous Environments (VLN-CE), which requires an agent to follow step-by-step natural language instructions to reach a target. We present a modular plan-and-control approach for the task. Our model consists of three modules: the candidate waypoints predictor (CWP), the history enhanced planner and the tryout controller. In each decision loop, CWP first predicts a set of candidate waypoints based on depth observations from multiple views. It can reduce the complexity of the action space and facilitate planning. Then, a history-enhanced planner is adopted to select one of the candidate waypoints as the subgoal. The planner additionally encodes historical memory to track the navigation progress, which is especially effective for long-horizon navigation. Finally, we propose a non-parametric heuristic controller named tryout to execute low-level actions to reach the planned subgoal. It is based on the trial-and-error mechanism which can help the agent to avoid obstacles and escape from getting stuck. All three modules work hierarchically until the agent stops. We further take several recent advances of Vision-and-Language Navigation (VLN) to improve the performance such as pretraining based on large-scale synthetic in-domain dataset, environment-level data augmentation and snapshot model ensemble. Our model won the RxR-Habitat Competition 2022, with 48\% and 90\% relative improvements over existing methods on NDTW and SR metrics respectively.
{\let\thefootnote\relax\footnote{{* Equal contributions}}}
\end{abstract}

\section{Introduction}
\label{sec:intro}
The goal of vision-and-language navigation (VLN)~\cite{anderson2018vision} is to learn an agent which can navigate by following natural language instructions. 
Two distinct scenarios have been proposed for VLN research: navigation on predefined graphs (VLN)~\cite{anderson2018vision} and navigation in continuous environments (VLN-CE)~\cite{krantz2020beyond}.
VLN-CE is more difficult due to the low-level action space (e.g., \textit{forward} 0.25 meters or \textit{turn} 30 degrees) compared to the high-level teleportation on a predefined graph in VLN.
In another word, more practical navigation challenges are exposed in VLN-CE such as how to avoid obstacles when controlling the agent to move.

The CVPR 2022 RxR-Habitat competition provides the community a neat VLN-CE benckmark on RxR dataset~\cite{anderson2020rxr} using Habitat Simulator~\cite{savva2019habitat}.
RxR is a multilingual large-scale VLN dataset, which provides \textbf{fine-grained instruction} annotations for \textbf{long trajectories}.
It requires the agent to follow the instruction step-by-step strictly, thus \textit{Normalized Dynamic Time Wrapping} (NDTW)~\cite{ilharco2019general} is the primary evaluation metric. NDTW measures the fidelity between the predicted and the ground-truth trajectories.

There are three primary challenges in RxR-Habitat Competition. 
\textbf{The first challenge} comes from the continuous action space, where the agent could only perform low-level controls (e.g., \textit{forward} 0.25 meters or \textit{turn} 30 degrees). 
Grounding instructions to these low-level actions will lead to poor performance as shown in~\cite{hong2022bridging,krantz2020beyond}.
It can lead to loss of state tracking of the agent because a sub-instruction may correspond to multiple low-level actions.
\textbf{The second challenge} comes from the RxR dataset. The long-trajectory instruction-following task requires accurate perception and monitoring of the navigation progress. The commonly used single-vector state representation is insufficient to address the problem.
\textbf{The third challenge} is that the agent can easily early stop or get stuck because sliding along obstacles on collision is forbidden in this competition.
Previous work~\cite{hong2022bridging} showed that transferring the controlling method from sliding-allowed to sliding-forbidden environments can lead to dramatic performance drop, because existing VLN controlling methods are unaware of collision let alone escaping from getting stuck, which will result in navigation failures.

To address the above challenges, we propose a modular plan-and-control model for the instruction-following task. 
First,  we adopt a candidate waypoints \textbf{predictor} at the beginning of each decision loop to bridge the differences between discrete and continuous environments following~\cite{hong2022bridging}.  
The predictor could predict a local adjacent navigation graph (each node represents a candidate waypoint), thus a variety of VLN models~\cite{qiao2022hop,he2021landmark,qi2021road,an2021neighbor,chen2021history} could be transferred to VLN-CE seamlessly.
Furthermore, we decouple the navigation into planning and control. 
The \textbf{planning} aims at determining a candidate waypoint as the subgoal, which is very similar to the high-level teleportation action in discrete VLN task.
We adopt a transformer-based history enhanced planner to capture the navigation progress and infer the subgoal.
Specifically, an additional history encoder is proposed to encode historical memory to track the navigation progress.
This mechanism has been shown effective for navigation, especially for the long-horizon navigation such as VLN task based on RxR dataset~\cite{chen2021history,pashevich2021episodic}. 
Moreover, we design a non-parametric \textbf{controller} named \textit{tryout} to execute low-level actions to reach the planned subgoal.
Regarding the no-sliding setting in the competition, \textit{tryout}  controller could detect collisions and help the agent escape from getting stuck based on a heuristic trial-and-error mechanism.
%

We employ three recent advances in VLN to further improve the performance. 
We first adopt in-domain vision-and-language pretraining for our transformer-based planner using large-scale synthetic instructions~\cite{wang2022less} (around 1 million RxR-like instruction-trajectory pairs).
Then, we adopt style-transfer-augmentation~\cite{li2022envedit} on RGB observations when online tunning our agent. 
Also, we adopt snapshot model ensemble~\cite{qin2021explore} when inferring navigation trajectories.
Extensive experiments demonstrate the effectiveness of our whole method and each of its designs. Our model won the RxR-Habitat competition 2022 and outperforms previous methods by a large margin, e.g., \textit{Normalized Dynamic Time Wrapping} (NDTW) increases from 37.39\% to 55.43\% and \textit{Success Rate} (SR) increases from 24.85\% to 45.82\%.

\begin{figure}[!tbp]
	\centering
	\includegraphics[width=0.9\linewidth]{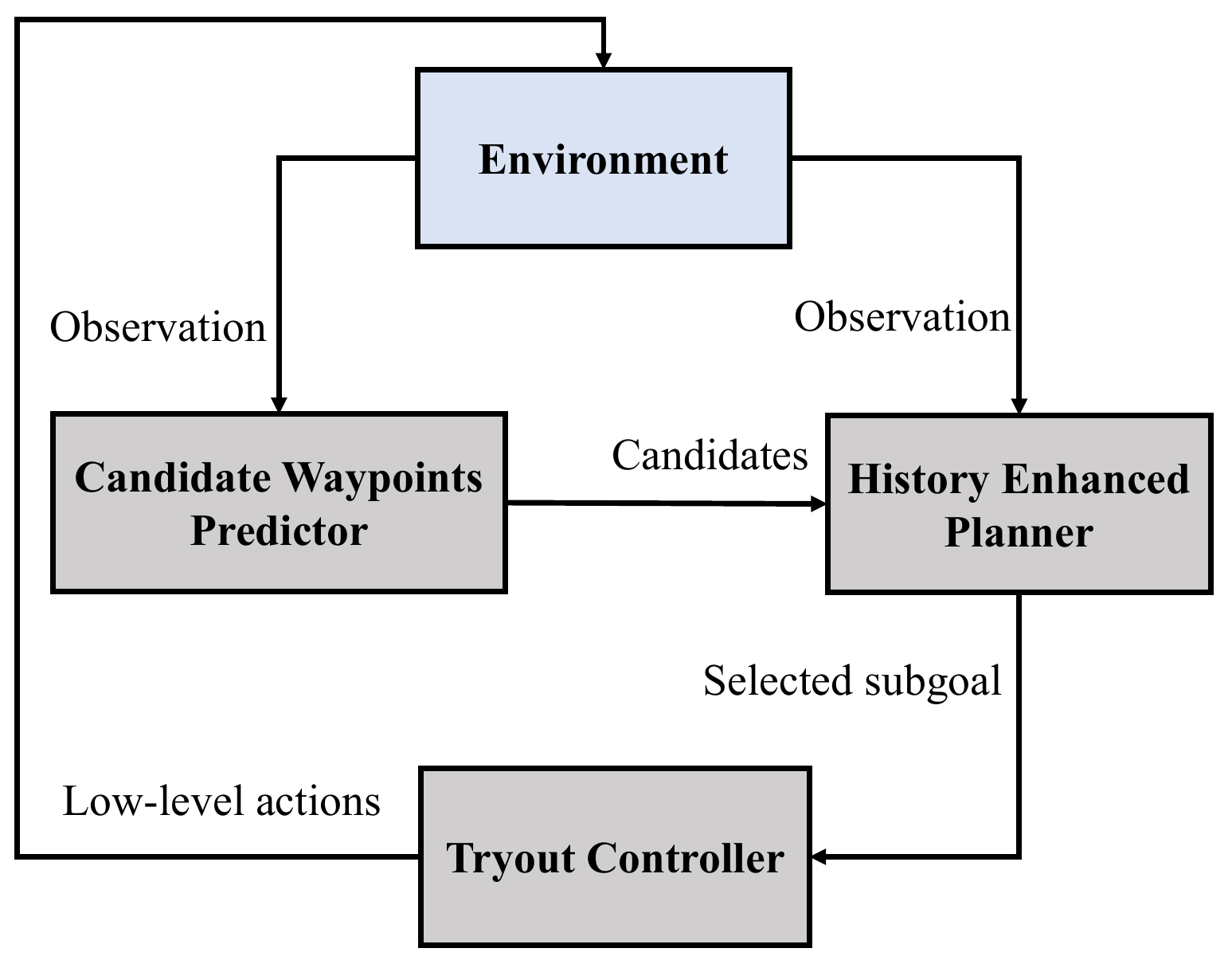}
	\vspace{-1mm}
	\caption{Overview of our modular plan-and-control scheme.}
	\label{fig:overview}
\end{figure}

\section{Method}
\label{sec:method}
Our proposed method consists of three modules (see Figure~\ref{fig:overview}): the candidate waypoints predictor, the history enhanced planner and the \textit{tryout} controller. 
In each decision loop, the predictor first predicts a set of candidate waypoints following~\cite{hong2022bridging}. 
Then, the history enhanced planner selects one of the candidate waypoints as the subgoal conditioned on the current and historical observations. 
Finally, the \textit{tryout} controller executes a sequence of low-level actions to reach the subgoal.
All three modules work hierarchically until the agent decides to stop. We present the three modules in detail below.


\subsection{Candidate Waypoints Predictor}
The candidate waypoints predictor predicts a set of candidate waypoints based on the current observations of the agent.
It results in a local adjacent navigation graph (each node represents a candidate waypoint) which is similar to the navigation graph in discrete VLN~\cite{anderson2018vision}. 
The predicted waypoints bridge the differences between discrete and continuous environments, and can result in efficient navigation as shown in~\cite{hong2022bridging}.
Moreover, a variety of advances in VLN could be transferred to VLN-CE seamlessly considering the narrowing environmental differences.
We utilize a slightly modified predictor in this work. The main differences between our predictor and the one in~\cite{hong2022bridging} lie in two aspects:
\begin{itemize}
    \item [1)] Both the two predictors take multiple views as input. Instead of using multiple sensors to get observations from multiple views, we rotate the agent to obtain multiple views. It follows the egocentric vision setting in RxR-Habitat 2022 strictly.
    \item [2)] Instead of taking both RGB and depth images as input, our predictor only takes depth images. We empirically find the predictor could achieve better generalizability.
\end{itemize}
Please note that the rotation actions in 1) are counted in the total action steps. 
Without loss of generality, some other recent methods~\cite{krantz2022sim,krantz2021waypoint} can also be adapted to predict waypoints in this framework.


\subsection{History Enhanced Planner}
Given a set of candidate waypoints, the planner aims at determining one of the waypoints as the subgoal to navigate. 
The planning task is almost the same as VLN in discrete environments, where the agent selects an adjacent node to teleport.
%
Different from VLN, the agent in VLN-CE moves to a subgoal by performing low-level actions instead of teleportation.
%
A variety of state-of-the-art VLN methods can be transferred to VLN-CE easily by taking advantages of predicted waypoints.
We take HAMT~\cite{chen2021history} as our base planner because of its advanced performance and compatibility to RxR-Habitat competition.
Specifically, HAMT adopts a history encoder to encode navigation memory in addition to typically used language and vision encoder.
%
The mechanism has been shown especially effective to long-horizon navigation task such as RxR navigation task~\cite{chen2021history,pashevich2021episodic}.
Furthermore, HAMT adopts a multilingual instruction encoder for RxR navigation. Thus, no repetitive training phases is necessary for different languages in RxR-Habitat which is computation-efficient.
Our history planner is slightly modified from HAMT by additional encoding depth images in vision encoder for better waypoints representations. We recommend referring to~\cite{chen2021history} for more details.


\subsection{Tryout Controller}
The controller drives the agent by a sequence of low-level actions to reach the subgoal selected by the planner. 
One of the most challenging settings in RxR-Habitat Competition is the \textbf{no-sliding} setting which means the agent is not allowed to slide along the obstacles on collision.
It simulates the real-world deployment scenario where the ability to avoid obstacles and deadlocks is necessary.
However, most of existing VLN-CE agents~\cite{krantz2020beyond,hong2022bridging,krantz2021waypoint} are unaware of collisions let alone escape from getting stuck, which can lead to navigation early stopping and failures.
To address this issue, we design a non-parametric heuristic controller named \textit{tryout} which could help the agent escape from deadlocks. 
Specifically, \textit{tryout} first transforms the selected subgoal (parameterized by the distance and heading) into a sequence of low-level actions. 
During the execution of these low-level actions, \textit{tryout} detects deadlocks by simply comparing the difference between the previous and the current frame. 
Once deadlock, the agent will try several predefined candidate directions with a single \textit{forward} action to escape.
The agent will proceed with low-level actions left if the escape is successful.
Note that both the executed and tried actions are counted in the total action steps. The agent would stop once the total steps exceed the max action steps.

\subsection{Training}
Our training phase contains two stages. The first stage is offline pretraining which aims at learning general visual-linguistic joint representation in a self-supervised manner.
We use the same proxy tasks with~\cite{chen2021history} such as \textit{Masked Language Modeling}, \textit{Instruction Trajectory Matching}, \textit{Single-step Action Prediction/Regression} and \textit{Spatial Relation Prediction}. 
The second stage is online fine-tuning, where we adopt schedule sampling used in~\cite{hong2022bridging} to tune our agent in Habitat simulator~\cite{savva2019habitat}.
Note that we train the agent with sliding allowed to avoid getting stuck and gain more experience, while evaluating our agent in the no-sliding setting with \textit{tryout}. 
We empirically found the agent learned more generalizable policy by using the sliding to no-sliding transfer scheme. 
%
%
In addition, our \textit{tryout} could greatly weaken the performance impact of the no-sliding setting when inferring trajectories.
Details will be discussed in Section~\ref{sec:ablation}.

\section{Experiments}
\label{sec:expr}
\subsection{Dataset and Evaluation Metrics}
The competition uses the RxR-Habitat dataset, which is a multilingual VLN-CE dataset transferred from the RxR dataset~\cite{anderson2020rxr} using the Habitat simulator. 
The scenes of RxR-Habitat are constructed from Matteport3D dataset~\cite{chang2017matterport3d}. 
It consists of 87K instruction-trajectory (IT) pairs in three languages (English (en), Hindi (hi), and Telugu (te)). 
The IT pairs are divided into train, val-seen, val-unseen, and test-unseen splits according to their corresponded scenes. 
More attention is paid to the performance in unseen environments, and the test-unseen split is used for the competition.
\textit{Normalized Dynamic Time Wrapping} (NDTW)~\cite{ilharco2019general} is the primary metric to evaluate the navigation performance of an agent since it measures the fidelity between the predicted and the ground-truth trajectories.
\textit{Success Rate} (SR) and \textit{Success Rate weighted by Length} (SPL) are the main secondary metrics.

\subsection{Implementation Details}
The Candidate Waypoints Predictor is trained using the same training setting and graph dataset following~\cite{hong2022bridging} with only depth features. 
We first pretrain the History Enhanced Planner in Matterport3D Simulator with  RGB images while the training data consists of RxR instruction-trajectory pairs and marky-mT5 pairs generated pairs from~\cite{wang2022less}. 
The model is trained for 300,000 iterations with a batch size of 48 and a learning rate of 5e-5 on 4 NVIDIA RTX 3090 GPUs. We then fine-tune the model in the Habitat simulator with RGBD images on RxR-Habitat dataset for 15,000 iterations using a batch size of 64 and a learning rate of 3e-5 via AdamW optimizer on 4 NVIDIA Tesla A100 GPUs.  The multilingual language encoder of the History Enhanced Planner is initialized from RoBERTa~\cite{liu2019roberta}, and the RGB images are encoded by ViT~\cite{dosovitskiy2020image} pretrained in CLIP~\cite{radford2021learning} while the depth images are encoded by ResNet50~\cite{he2016deep} pretrained in DDPPO~\cite{wijmans2019dd}.
We follow the standard competition configuration\footnote{\url{https://github.com/jacobkrantz/VLN-CE/blob/master/habitat_extensions/config}} and all models are trained using the train split.

\begin{table}[t]
	\centering
	\resizebox{0.48\textwidth}{!}{\begin{tabular}{lccccccc}
			\toprule
			\multicolumn{1}{c}{} & \multicolumn{3}{c}{Component} & \multicolumn{4}{c}{RxR-CE Val-Unseen} \\
            \cmidrule(r){2-4} \cmidrule(r){5-8} 
			\multicolumn{1}{c}{model} &
			\multicolumn{1}{c}{mT5-P} & \multicolumn{1}{c}{ST} & \multicolumn{1}{c}{En} & 
			\multicolumn{1}{c}{PL}  & \multicolumn{1}{c}{SR$\uparrow$} & \multicolumn{1}{c}{SPL$\uparrow$} & \multicolumn{1}{c}{NDTW$\uparrow$}\\
			\midrule
			1 & & & & 14.49 & 47.56 & 41.65 & 63.02 \\
			2 & \checkmark & & & 14.1 & 48.6 & 42.05 & 63.35 \\
			3 & \checkmark & \checkmark & & 13.93 & 49.33 & 43.4 & 64.06 \\
			4 & \checkmark & \checkmark & \checkmark & 13.97 & 49.6 & 43.4 & 64.64 \\
			\bottomrule
	\end{tabular}}
	\caption{Ablation studies of training and inference strategies on val-unseen (English) split, where mT5-P denotes joint pretraining with RxR and marky-mT5, ST denotes style transfer as environment-level augmentation, and En denotes different snapshots ensemble.}
	\label{tab:abalation}
\end{table}

\begin{table}[t]
	\centering
	\resizebox{0.48\textwidth}{!}{\begin{tabular}{lcccccc}
			\toprule
			\multicolumn{1}{c}{} & \multicolumn{2}{c}{Sliding} & \multicolumn{1}{c}{} & \multicolumn{3}{c}{RxR-CE Val-Unseen} \\
            \cmidrule(r){2-3} \cmidrule(r){5-7} 
			\multicolumn{1}{c}{model} &
			\multicolumn{1}{c}{Train} & \multicolumn{1}{c}{Infer} & 
			\multicolumn{1}{c}{Tryout} & 
			\multicolumn{1}{c}{SR$\uparrow$} & \multicolumn{1}{c}{SPL$\uparrow$} & \multicolumn{1}{c}{NDTW$\uparrow$}\\
			\midrule
			1 & \checkmark & & & 20.93 & 19.31 & 44.91 \\
			2 & \checkmark & \checkmark & & 51.98 & 46.13 & 64.95 \\
			\midrule
			3 & & & \checkmark & 45.0 & 38.55 & 59.97 \\
			4 & \checkmark & & \checkmark & 48.6 & 42.05 & 63.35 \\
			\bottomrule
	\end{tabular}}
	\caption{The influence of sliding and \textit{tryout} controller.}
	\label{tab:sliding}
\end{table}


\subsection{Ablation Study}
\label{sec:ablation}
\noindent \textbf{Analysis of training and inference strategies.}
Table~\ref{tab:abalation} shows the ablation study about our training and inference strategies. 
Note that all models are trained in sliding-allowed train split and evaluated in sliding-forbidden val-unseen split with \textit{tryout} controller.
Model\#1 in Table~\ref{tab:abalation} denotes our base planner which is initialized by History Enhanced Planner pretrained with RxR dataset only.
It achieves 63.02\% NDTW and 41.65\% SR, which has already improved previous arts~\cite{hong2022bridging} by a large margin.
By pretraining with RxR and marky-mT5 mixed instruction-trajectory pairs, model\#2 further improves our base planner by 0.3\% NDTW and 1\% SR. 
Since marky-mT5 has an order of magnitude more pairs compared to RxR (1M vs. 0.1M), better linguistic-visual joint representations can be learned. 
Inspired by~\cite{li2022envedit}, we employ style transfer as environment-level augmentation during online tuning of our agent (see model \#3). It gains 0.7\% more NDTW and SR by learning style-invariant visual representation. 
Finally we adopt snapshot model ensemble~\cite{qin2021explore} in inference to further improve the performance. We simply ensemble models from two different training iterations as the planner. It achieves 0.6\% more NDTW  and 1.3\% more SR (see model\#4).

\vspace{1mm}
\noindent \textbf{Analysis of the no-sliding setting.} 
Table~\ref{tab:sliding} demonstrate the influence of the sliding and our \textit{tryout} controller. This ablation experiment is based on model\#2 in Table~\ref{tab:abalation}.
The upper bound of the performance corresponds to model\#2 in Table~\ref{tab:sliding}, where sliding is allowed both in training and inference.
The navigation performance dramatically drops from 51.98\% SR to 20.93\% SR (see model\#1) when the sliding is turned off in inference.
It indicates the severe impact of the no-sliding setting on navigation performance and similar phenomenon is also observed in~\cite{hong2022bridging}.
The reason is that the no-sliding setting leads to early-stopping and getting stuck, which results in navigation failures.
However, the navigation performance can be recovered from 20.93\% SR to 48.6\% SR by taking advantage of our \textit{tryout} controller (see model\#4). This indicates \textit{tryout} can bridge the performance gap between sliding-allowed and sliding-forbidden environments to a certain extent.
Furthermore, we turn off sliding in both training and inference phases in model\#3. A clear performance gap is observed comparing model\#3 to model\#4. The potential reason is that more experience and better instruction-trajectory alignments can be earned by training the agent in sliding-allowed environments.

\begin{table}[t]
	\centering
	\resizebox{0.48\textwidth}{!}{\begin{tabular}{lrrrrrr}
			\toprule 
			\multicolumn{1}{c}{} & \multicolumn{6}{c}{RxR-CE Test-Unseen} \\
			\cmidrule(r){2-7}
			\multicolumn{1}{c}{model} & 
			\multicolumn{1}{c}{PL} & \multicolumn{1}{c}{NE$\downarrow$} &
		    \multicolumn{1}{c}{SR$\uparrow$} & \multicolumn{1}{c}{SPL$\uparrow$} &
		    \multicolumn{1}{c}{NDTW$\uparrow$} & \multicolumn{1}{c}{SDTW$\uparrow$} \\
			\midrule
			VLN-CE & 7.33 & 12.1 & 13.93 & 11.96 & 30.86 & 11.01 \\
			CWP-CMA & 20.04 & 10.4 & 24.08 & 19.07 & 37.39 & 18.65 \\
			CWP-RecBert & 20.09 & 10.4 & 24.85 & 19.61 & 37.30 & 19.05 \\
			\midrule
			Reborn (Ours) & 16.0 & \textbf{7.1} & \textbf{45.82} & \textbf{38.82} & \textbf{55.43} & \textbf{38.42} \\
			\bottomrule
	\end{tabular}}
	\caption{RxR-Habitat 2022 leaderboard results.}
	\label{tab:leaderboard}
\end{table}

\subsection{Leaderboard Results}
Table~\ref{tab:leaderboard} shows our final results in the RxR-Habitat competition 2022\footnote{\url{https://ai.google.com/research/rxr/habitat}}. Our model (Reborn) significantly outperforms previous state-of-the-arts through all evaluation metrics, e.g., SR increases from 24.85\% to 45.82\% and NDTW increases from 37.39\% to 55.43\%. Note that our submission is inferred by model\#2 in Table~\ref{tab:abalation} due to the submission time restriction.

\section{Conclusion}
\label{sec:Conlusion}
The RxR-Habitat Competition introduces a challenging VLN task with a harder dataset (RxR) and a harder simulator (Habitat) compared with R2R in Matterport3D simulator. 
By combining several recent advances in VLN, our method significantly improves over existing baselines, i.e., the NDTW metric increases from 37.39\% to 55.43\%, and the SR metric increases from 24.85\% to 45.82\%.
However, the performance is still far from perfect. We hope that our method provides a strong baseline for further research.


{\small
\bibliographystyle{ieee_fullname}
\bibliography{egbib}
}

\end{document}